# The Expected Unexpected & Unexpected Unexpected:
## How People's Conception of the Unexpected is Not Really That Unexpected

**Molly S. Quinn** (molly.quinn@ucdconnect.ie)
**Kathleen Campbell** (kathleen.campbell@ucdconnect.ie)
**Mark T. Keane** (mark.keane@ucd.ie)
**School of Computer Science & VistaMilk SFI Research Centre,**
**University College Dublin, Belfield, Dublin 4, Ireland**

**Abstract**

The answers people give when asked to "think of the unexpected" for everyday event scenarios appear to be more expected than unexpected. There are *expected unexpected* outcomes that closely adhere to the given information in a scenario, based on familiar disruptions and common plan-failures. There are also *unexpected unexpected* outcomes that are more inventive, that depart from given information, adding new concepts/actions. However, people seem to tend to conceive of the unexpected as the former more than the latter. Study 1 tests these proposals by analysing the object-concepts people mention in their reports of the unexpected and the agreement between their answers. Study 2 shows that object-choices are weakly influenced by recency, that is, the order of sentences in the scenario. The implications of these results for ideas in philosophy, psychology and computing are discussed.

*As we know, there are known knowns; there are things that we know that we know. We also know there are known unknowns; that is to say we know there are some things we do not know. But there are also unknown unknowns, the ones we don't know we don't know.*

Donald Rumsfeld, Feb 2002, US Secretary of Defence

## 1. Introduction

In an uncertain and contingent world, our ability to deal with the unexpected often gives us safe passage through the Siren-like obstacles of everyday life (see e.g., Weiner, 1985a). The Cognitive Sciences have often concerned themselves with how people think about the unexpected. However, most of this research relies on theory-driven definitions of the unexpected (e.g., low probability events), rather than simply asking people to "think of the unexpected" and see how they respond[1]. In the present paper, we report two studies that ask people to generate unexpected events for everyday scenarios and then analyse their responses. As the title of the paper suggests, our main finding is that *people's conception of the unexpected is not really that unexpected at all.*

In Cognitive Psychology, *unexpectedness* is often used as a dependent variable in studies of human thinking and decision making. For example, in reasoning research, the unexpected has often been proposed to elicit counterfactual thinking (Kahneman & Miller, 1986; McEleney & Byrne, 2006). In attribution research, the unexpected has been cast as non-normative behavior in others, that elicits spontaneous causal thinking (Hastie, 1984; Weiner, 1985b). And, surprising events are often defined in terms of their unexpectedness (Meyer et al., 1997; Maguire et al., 2011).

However, most of these studies do not actually *ask* people to report unexpected outcomes; rather they adopt *a priori* operational definitions of *unexpectedness* based on the experimenter's theoretical stance. The unexpected is commonly operationalized as (i) an event rated as having a low subjective probability (Maguire et al., 2011; Teigen & Keren, 2003), or (ii) profiles of people with inconsistent traits (Hastie, 1984), or (iii) events that are simply asserted to be unexpected to the actors in a narrative (McEleney & Byrne, 2006). In contrast, we do not use *a priori* definitions but rather, simply, ask people to tell us what they consider the unexpected to be. This sort of behaviour was observed by Foster & Keane (2015, 2019), in studies on surprise, in the form of familiar surprises ("I am surprised my wallet is missing from my trouser pocket, but I am guessing it was robbed") and unfamiliar surprises ( "I am surprise my belt is missing from my trousers but have no idea how that could have happened"; see also Maguire & Keane, 2006).

### 1.1 Thinking About the Unexpected

Consider the simple task used in the current experiments to elicit unexpected events from people. Imagine being told a story about a woman, called Louise, who is going shopping at her favourite clothes store, in which she draws money from an ATM and heads into town on the bus. Now imagine you are told "Something unexpected occurred. What do you think happened?" One could respond with one of the following unexpected events, saying that:

1) Louise lost the money she drew from the ATM.
2) Louise was delayed in traffic, arrived late and the shop was shut.

However, one could also validly say:

3) The bus stopped at a charity bus-wash and Louise got covered in suds.
4) Louise pulled a gun on the driver and robbed him to raise more money for her shopping spree.

Intuitively, as unexpected outcomes, the first two answers (1-2) are quite conservative and mundane and *less* unexpected than the latter two responses (3-4) which are more inventive and a lot *more* unexpected. We call the former answers the *expected unexpected* and gloss the latter as the *unexpected unexpected*.

---
[1] Khemlani et al.'s (2011) Expt. 3 is a notable, but rare, exception though it focusses on the issue of latent scope.

*Expected unexpected* outcomes tend to maintain the original goal of the story scenario (i.e., shopping) and the stated object-concepts associated with the story's goal; these *goal concepts* tend to be re-used in the unexpected event (i.e., *bus*, *ATM*, *money*, *store*) and few new objects are added (e.g., *traffic*). Furthermore, these events are often "common failures" that are familiar to people; losing one's money or being delayed are common reasons for failed plans and goals.

*Unexpected unexpected* outcomes, in contrast, may establish new goals for the story scenario (e.g., attending a charity event) and, though *goal objects* may be used (i.e., *bus*, *money*), often "new" object-concepts not present in the original story are introduced (e.g., *guns, suds*). Also, these unexpected events are quite unfamiliar to the scenario: getting involved in a charity bus-wash is not a common everyday event for most people who are going shopping. In previous work on surprise using these everyday scenarios (Foster & Keane, 2015), we noticed that people typically produced *expected unexpected* answers rather than *unexpected unexpected* ones. But, why?

Why do people *minimally* perturb the stated scenario, keeping its goals and goal-concepts in these expected-unexpected events that they seem to prefer to generate? One possibility is that when people are thinking about the unexpected, they are essentially trying to explain how current goals might fail; so, unexpected events tend to describe disruptions to a current plan or undoings of assumed facts that enable current goals. Being delayed in traffic disrupts Louise's shopping plan, undoing the assumption that the bus gets her to the shop on time. Losing one's money is an unexpected event that explains how any shopping-goal might fail. From an adaptive perspective, it makes sense to minimally change the current situation when projecting such unexpected futures. In contrast, more creative unexpected-unexpected events, that depart significantly from the current scenario, may never occur and, therefore, seem not to be considered. In short, the former probably have higher predictive value than the latter. Across the Cognitive Sciences, many researchers have highlighted this minimalist stance in people when they encounter the unexpected.

**1.2 The Minimalism of the Unexpected**

In Philosophy, when reasoning about inconsistencies (such as new, unexpected facts), it has been repeatedly proposed that any change to stated propositions or prior beliefs should be as minimal as possible; observing the "maxim of minimal mutilation" (Quine, 1992, p.14), or the "principle of conservativism" (Harman, 1986, p.46). Similarly, in considering counterfactual situations (of which unexpected situations could be a subclass), Lewis (1986), taking a possible-world perspective, talks of finding the maximally-similar world to the current one.

In Psychology, related ideas arise in considering the *minimal-mutability* of counterfactual scenarios (Kahneman & Miller, 1986). Also, in the psychology of explanation, several researchers have noted how explanations of the unexpected maintain aspects of the original scenario; they preserve the level-of-abstraction of the original scenario rather than identifying new or more specific information (see e.g., Johnson & Keil, 2014) or they favor explanations with a narrow, latent scope (Khemlani et al., 2011).

In Artificial Intelligence, theories of understanding and explanation directly predict *minimalism* and show how the "expected unexpected" might arise (see Leake, 1991, 1992; Schank, 1986; Schank, Kass & Riesbeck, 1994). David Leake's (1992) computational account of understanding gives the most comprehensive account of what people might be doing when asked to "think of the unexpected" (see also Schank, 1986). Leake argues that people store *explanation patterns* to handle plan failures and anomalies encountered in everyday life. These explanation patterns can be thought of as "script-like" structures (Schank & Abelson, 1977), at varying degrees of abstraction, that can account for difficulties that arise in plans; for instance, in considering how a planned shopping-expedition might be disrupted, a number of standard disruption-events suggest themselves from pre-stored explanation patterns (e.g., that I might be mugged, or that I might lose my money or that I might be delayed). To handle a contingent world, it is proposed that we store these pre-canned explanations and retrieve them to quickly explain unexpected happenings[2]. Although these ideas have been referenced in the psychological literature (e.g., Hastie, 1984), they have not been worked up into a psychological model or specifically tested. Here, we propose an initial psychological account, that we then test this model in two experiments.

**1.3 Minimal Retrieval Model**

Our psychological account for the generating unexpected events is called the Minimal Retrieval Model (MRM). According to this model, when people are asked for unexpected outcomes to everyday scenarios, they retrieve explanation patterns and adapt them to the situation in hand. Specifically, that people build a *cue frame* using the given information in the scenario (e.g., the goals, actors, actions and objects mentioned) to search memory for suitable explanation patterns. For example, when people are told Louise had the goal of going shopping, took money from the ATM and then went to town, it is assumed that memory is searched for explanation patterns involving shopping-goals, female-shoppers, buses, money, and ATMs. Accordingly, unexpected events such as Louise losing her money, having problems with the ATM, or being delayed will tend to be found in memory and returned as responses, rather than more inventive answers.

Minimal Retrieval Model makes several predictions about the nature of the unexpected outcomes reported by people; specifically, it is predicted that (i) reported unexpected events should tend to use the stated object-concepts in the original scenario because memory will be *cued* with these concepts and the retrieved explanation patterns will instantiate these objects, (ii) goal-related objects will be preferred in reported events, over non-goal objects, (iii) people will tend to agree on the reported unexpected events because they are using familiar plan-failures (i.e., explanation patterns). Note, the first two of these predictions basically propose that minimalism is a side-effect of the retrieval process and the third prediction basically says that answers will be expected-unexpected events rather than unexpected-unexpected ones.

---

[2] Note, both Leake and Schank maintain that not all situations can be handled by these pre-canned explanation-patterns; explaining the unexpected may sometimes involve much more creative uses of prior knowledge, such as analogical explanations.

Table 1: Louise-Shopping Story & Answer Categories

*Sentence Order Used in Study 1 & 2 (Normal Condition)*

| | | |
|---|---|---|
| Goal | (S1) | Louise wants to shop at an expensive clothes store. |
| Non-Goal | (S2) | She is wearing her favourite dress and matching shoes. |
| Goal Step | (S3) | Louise draws money from the ATM. |

*Sentence Order Used in Study 2 (Reversed Condition)*

| | | |
|---|---|---|
| Goal | (S1) | Louise wants to shop at an expensive clothes store. |
| Goal Step | (S2) | Louise draws money from the ATM. |
| Non-Goal | (S3) | She is wearing her favourite dress and matching shoes. |

*Answer Categories for Unexpected Events*

| | |
|---|---|
| *ls_neg_ans1* | She has insufficient money to buy |
| *ls_neg_ans2* | She has problems with the ATM |
| *ls_neg_ans3* | She is robbed or loses money/card/id. |
| *ls_neg_ans4* | Clothes issues (dress rips, shoe snaps). |
| *ls_neg_ans5* | The shop is closed. |
| *ls_pos_ans1* | She finds or ATM gives more money. |
| *ls_pos_ans2* | She has more money than she thought. |
| *ls_pos_ans3* | Good events involving shoes and dress. |
| *ls_pos_ans4* | Sale is on at the shop. |
| *ls_other* | e.g.; ATM speaks, gives money to charity. |

**Preference for Stated Object-Concepts**. If memory is being searched with the stated goals and object-concepts given in the scenario then the explanation patterns retrieved should reflect these objects/entities and minimally introduce new objects (e.g., ones that mention *money*, *ATMs*, *buses*). This process thus delivers unexpected events that remain close to the original scenario, with perhaps better predictive value. In Quine's terms, the reported unexpected event will *minimally mutilate* the original scenario. In the present studies, we measure this minimalism by recording the frequency of stated objects versus new objects in the reported unexpected outcomes (excluding references to *Louise* who as the main actor will always tend to be mentioned).

**Preference for Goal Objects**. Within the preference for stated objects, MRM also predicts that goal-related objects will be preferred over less goal-related objects (which we will call *non-goal objects*). For example, if the scenario mentions that "Louise was wearing her favorite dress and matching shoes" (see Table 1), these objects *dress* and *shoes* are less goal-critical. People need *money* to go shopping but what they wear is less critical to the shopping goal [3]. Although, it is feasible to generate unexpected events from these non-goal objects (e.g., "when Louise got on the bus, everyone was wearing the same dress and shoes"), explanation patterns based on non-goal objects are less likely to be retrieved because they are not goal-critical. In the present studies, we check for this preference by recording the frequency of stated goals-objects versus non-goal objects in the reported unexpected outcomes (obviously, again, excluding references to *Louise* who as the main actor will tend to be mentioned anyway).

**Agreement**. By definition, explanation patterns are explanations for commonly-occurring disruptions to everyday plan-goal sequences; it makes more sense for the cognitive system to assume that disruptions that happened repeatedly in the past will happen again. As such, they should be familiar to people, they should be expected-unexpected events. Hence, there should be a high level of agreement between people in the unexpected outcomes they propose. This means that most answers should fall into a small set of common answer-categories; for instance, we should see many people using answers that describe "Louise losing her money" or "the shop being shut" (see Table 1). In the present studies, we test this prediction by classifying people's responses into answer-categories and recording the proportion of answers that fall into these categories. In the remainder of this paper, we report two experiments designed to test these predictions. To the best of our knowledge, these tests are new, as are the measures used to assess what people report as the unexpected.

## 2. Study 1: How Unexpected?

The study presented participants with scenarios describing everyday events such as going shopping, doing exams, going on trips and attending business meetings (adapted from Foster & Keane, 2015). Each story was followed by an instruction to think of the unexpected. The unexpected outcomes reported for each scenario were categorized by three judges in terms of object-concepts (i.e., *goal-objects*, *non-goal-objects*, *both* goal- and non-goal-objects and *neither* of the stated objects) and answer-categories used.

---

[3] Though, obviously, it could be made goal-critical with additional conditions (e.g., if one said "she wanted to be able to match the clothes she was wearing with those in the shop").

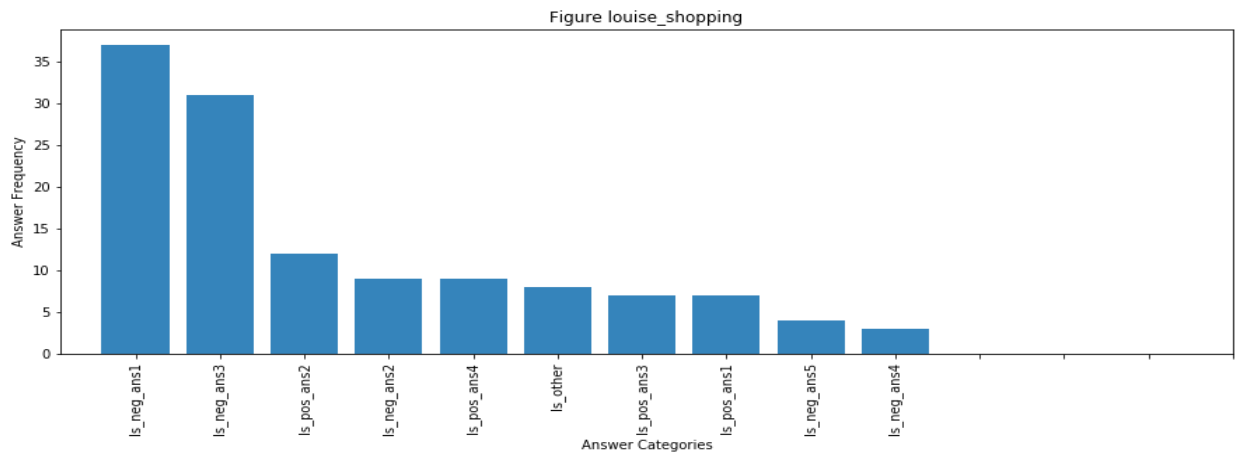

*Fig 1. Frequencies for Answer Categories in Louise Story (see Table 1 for Meaning of Answer Code)*

### 2.1 Method

**Participants & Design.** The study involved 127 participants and was run on the crowdsourcing platform, prolific.com[4]. Participants were native English speakers from Ireland/UK/USA and had not participated in previous studies by the group.

**Procedure & Materials.** All participants received the same 20 scenarios (randomly re-ordered for each participant), after first being shown two practice materials (these items were not flagged as practice items and not included in the analysis). Participants were presented with a series of web pages, explaining the task and then presented with one scenario after the other, each on its own page. Each material was presented along with the associated instruction to think of unexpected outcomes to the scenario (material lists available on request).

Each scenario was described in three sentences: a setting goal-sentence (S1; Louise going shopping), one giving additional information that was not on the critical path of the plan for the goal (S2; Louise wearing her favourite dress and shoes) and a final one describing some further action taken to achieve the goal (S3: Louise drawing money, see Table 1). Two comprehension questions were asked about the scenario, as a test micro-task, to ensure participants had carefully read and understood the scenario. The instruction to think of the unexpected followed these questions. Participants wrote their responses in a text-field with no upper limit.

**Measures & Judgements.** In total the study yielded 2,540 responses (127 participants x 20 materials) each of which were judged by three raters (i.e., the three authors) for answer-type and the use of goal-related and non-goal-related objects (the main actor of the scenario was always excluded from this object judgement). Every response was categorized into answer categories specific to the material (e.g., in Louise scenario, answers about losing money, being delayed, the shop being shut). The classification of answers by the three judges revealed high levels of agreement: pairwise comparisons between judgements revealed Cohen's Kappas from K=0.82 to K=0.89. The classification of objects in the answers (as goal, non-goal, both, neither stated object) by the three judges had lower, but acceptable, levels of agreement. Agreement between Judge-1 and Judge-3 was lower (K=0.56) as object classifications were re-defined more tightly for Judge-2 and 3, who agreed more often (K=0.74). The final classifications chosen for all judgements was based on a majority vote from the three judges. Three-way splits (which were rare, N<20) were resolved by discussion.

### 2.2 Results & Discussion

Overall, the results confirm the predictions made from the Minimal Retrieval Model; the unexpected is really not that unexpected. People tend to (i) stick to the stated objects in the scenario rather than use new objects, (ii) they show a strong preference for given goal-objects over non-goal objects, (iii) they agree on the unexpected events reported, as a few answer-categories cover most responses made.

**Preference for Stated Objects.** As predicted, people tend to stick to the object-concepts given in the scenario (e.g., the *money*, *buses*, *shoes* of the Louise story), rather than introducing new objects into their answers. Of the 2,540 unexpected outcomes reported by participants, 78% (N=1891) relied on the given objects, while only 22% (N = 649) of answers mentioned none of the stated objects (i.e., "Louise met her best *friend*"). Most unexpected outcomes assert a new relation between the given objects (e.g., "The *ATM* showed *Louise* had more *money*").

**Preference for Goal over Non-Goal Objects.** Furthermore, of the 78% (N=1,891) of unexpected outcomes that used the given objects from the scenario, the majority used only goal-objects (80%; N=1,518) with a minority using the non-goal objects (14%; N=261) and a few using both stated object-types (6%; N=112). Chi$^2$ tests performed on frequencies of the four object-types (df=3) for each material were all statistically significant at $p < 0.01$. Also, a by-materials analysis, using Wilcoxen's test, revealed a statistically significant difference in the proportions of goal versus non-goal objects chosen, $z = 3.00, p < .001$.

---

[4] The original experiment was divided into four conditions that used variants on the main instruction: asking for "something unexpected", "something good and unexpected", "something bad and unexpected", or "what would happen if the goal failed". For brevity, this manipulation is not reported here, as the same pattern of responding is seen across all these four conditions.

Of course, one could argue that this result is not surprising, as two sentences mentioned goal-objects (S1 and S3) and only one mentions non-goal objects (S2, see e.g. Table 1)[5]. However, even if examine goal and non-goal object choices at the sentence level, the preference for goal-objects remains: on average, goal-objects are chosen from S1 (M=27%) and S3 (M=44%) more often than the non-goal-objects from S2 (M=13%; see Table 2). Chi$^2$ tests performed on 40 pairwise comparisons of choices, for S1xS2 and S2xS3, found that only 5 comparisons were non-significant (most are $p < .001$). However, it is clear that there is a preference for goal-objects from the final sentence in the scenario (S3 at 44%), suggesting a recency effect, that we explore in Study 2.

Thus far, the evidence suggests that people respond in a minimalist way, sticking close to the original scenario's objects, with a strong preference for stated goal-related objects over stated non-goal objects.

**Agreement in Answer Categories.** Apart from analysing the object-concepts used in the unexpected outcome, we also categorised responses and noted their frequency of occurrence (e.g. see Table 2 and Fig.1). On average, materials were found to involve 10 answer categories (M=10.68, SD=1.7); Min=6 (*lucy_loan*) and a Max=13 (*robert_essay*; see Table 2). Figure 1 shows a typical distribution of responses across answer-categories for the Louise story; note, the top-3 most-frequently-used answer-categories of 10 categories tend to cover most responses (63%) followed by a long tail of lower-frequencies for other categories. Note, one answer-category was used as a residual one (the *other* category), and it typically also has a low count.

Table 2 shows the percentage of responses that fall into the top-3 most-used answer-categories for each material. In the most extreme case, *lucy_loan*, 89% of responses are covered by the top-3 answer-categories, with the lowest being 49% (for *bill_holiday*). This pattern of responding shows that there are very high levels of agreement between people with respect to the unexpected events they propose. For example, in the *louise_shopping* scenario, 29% of people proposed that Louise had money problems such as spending too much or not having enough money for the clothes (*ls_neg_ans1*), 25% proposed that she lost her money in some way (*ls_neg_ans3*) and 9% said that the ATM told her she had more money than she thought (*ls_pos_ans2*). None of these unexpected events are particularly "unexpected"; they are rather, typical disruptions that occur in everyday plans to achieve mundane goals. They are *expected unexpected* events. Indeed, more inventive answers -- the *unexpected unexpected* -- are quite rare and typically found in the *other* category. For instance, in the Louise story, the *other* category (*ls_other*, N=8) includes responses about (i) Louise deciding to give her money to a charity instead, (ii) Louise being approached by a film director who says she is beautiful and wants to make her a star and, (iii) the wonderful "The ATM opens, and Louise realizes it is a portal to her happiest childhood memory". These sorts of answers are the *unexpected unexpected*, truly unusual possible outcomes but, notably, are rare too.

### 3. Study 2: The Recently Unexpected?

Study 1 supports the minimalist predictions that people will stick closely to the original scenario, introduce few "new" objects and agree with others when proposing unexpected events for everyday scenarios. However, with respect to the object-concept analyses, the preference for goal-objects (especially, objects from the final sentence, S3) and lack-of-preference for non-goal objects could be due, in part, to a recency effect. That is, maybe people follow on from the last sentence in the story and, hence, use its goal-objects. For example, in the Louise story people do not mention her *shoes* and *dress* (the non-goal objects from S2) but rather follow on from the mention of *ATMs* and *money* (from S3) in proposing their unexpected outcome. If this were true then people's object-choices perhaps hinge less on their goal or non-goal status but more on order of mention. In this study, we put the non-goal sentence last (S3) to check if this changes the object-choices made (see Table 1 for a sample material).

### 3.1 Method

**Participants and Design.** The study involved 258 participants on the prolific.com crowdsourcing platform[5,6]. All were native English speakers from Ireland/UK/USA and had not taken part in our previous studies.

**Procedure & Materials.** All participants received the 20 scenarios used in Study 1, using the same procedure. There were two main conditions of interest: Normal (N = 126) and Reversed conditions (N = 132). Participants in the Normal condition received the same materials as those used in Study 1. Participants in the Reversed condition received variants of these materials, in which the non-goal sentence was moved to the last position in the story (S3; see example in Table 1).

**Measures & Judgements.** In total the study yielded 5,160 responses (258 participants x 20 materials). Given the very large number of responses in this experiment, we automated the object-judgement process (program and data will be made available on email request). A program, called ObjJudge was developed using the NLTK, Pandas and SciPy python packages to process the answers and identify whether they mentioned goal-objects, non-goal-objects, both object-types or neither. The program was trained on responses and their respective judgments from Study 1. Using lists of the object-entities (i.e.; object words given in the scenarios and synonyms provided in responses from Study 1), ObjJudge sorts the responses given in Study 2 into goal-object, non-goal object, both, or neither categories.

---

[5] Also, note the sentences themselves in the original scenarios mentioned equivalent numbers of objects (by-materials, paired t-tests on the object counts in S1, S2 and S3 revealed no differences, all $ps > 0.10$).

[6] The original experiment had 6 conditions that used (i) the three instructional variants used in Study 1 ("unexpected", "good and unexpected", "bad and unexpected"), crossed with (ii) a variant to think of "bizarre" events. Initial, analyses suggested that these variables do not impact the pattern of responding for object choices and, for brevity, are not reported here.

Stated simply, this program matches object-entities in the response-string against object-lists for each material (including common synonyms that people used in Study 1 answers). With this program, we achieved a high accuracy over all materials comparing its object-judgements against the human-judgments from Study 1 (M=93%; Min=90%, Max = 97%; Cohen's Kappa was K=1). In all other respects, the object-judgement measures were as detailed in Study 1.

### 3.2 Results & Discussion

The results replicate the findings of Study 1 that people (i) stick to the stated objects in the scenario rather than using new objects, (ii) they show a strong preference for the given goal-objects over non-goal objects. However, it also shows that here is a slight recency effect, in the Reversed condition, where the choice of non-goal objects increased by about 9% relative to the Normal condition.

**Preference for Stated Objects.** As we saw in Study 1, people tend to stick to the object-concepts given in the scenario (e.g., the *money*, *shoes* of the Louise story), rather than introducing new objects into their answers (e.g., *guns*, *suds*). Of the 5,160 unexpected outcomes reported by participants, 79% (N=4,098) made use of objects stated in the scenario, while only 21% (N = 1,062) of answers mentioned none of the given objects Most of the unexpected outcomes reported created a new relation between the given objects (e.g., "The *ATM* showed *Louise* had more *money*").

**Preference for Goal over Non-Goal Objects.** In a similar vein, of the 4,098 (79% of 5,160) unexpected outcomes that used objects from the original scenario, the majority mentioned only goal-objects (56%; N=2,869) whereas a minority mentioned only non-goal objects (13%; N=688), with some responses mentioning both goal and non-goal objects (10%; N=541). Chi$^2$ tests performed on frequencies of object-types reported for each material were all significant at $p < 0.05$ (df=3, with corrections for low-valued cells).

However, these analyses collapse across the Normal-Reversed manipulation designed to test for recency. When these conditions are broken out there is a small but statistically-significant increase in the use of the non-goal objects (roughly 9%, with a corresponding reduction in goal-object choices). The following are the relative percentages, for each choice category, Chi$^2$(3) = 76.9, $p < 0.001$:

| *Cond.* | *Goal* | *Non-Goal* | *Both* | *Neither* |
|---|---|---|---|---|
| *Normal* | 61% | 10% | 9% | 20% |
| *Reversed* | 51% | 16% | 12% | 21% |

In short, when the sentence with the non-goal objects is last in the story, people prefer to use the non-goal objects somewhat more often; showing that they are sensitive, to some degree, to the order in which information is given, though the dominance of goal-object choice still remains.

## 4. Conclusions

To the best of our knowledge, the current study is the only one simply asking people to "think of the unexpected" when presented with everyday scenarios. Our view is that when people are asked to do this, they tend to recall characteristic explanation patterns that account for common disruptions to everyday plans and goals (e.g., losing a resource, being delayed in executing a plan step). Accordingly, people report unexpected events are not really that unexpected; the *expected unexpected*. This work shows that these are a class of unexpected events -- things that commonly go wrong – that are to be distinguished from "truly unexpected" events (as Foster & Keane, 2015, found for surprising events). These findings should prompt a re-assessment of what we mean by the "unexpected" as a dependent variable in exploring aspects of human thought. It also raises the interesting prospect, that there is a lot more to be discovered about what people conceive the unexpected to be.

## Acknowledgements

The first author was supported by a scholarship from the School of Computer Science, University College Dublin, Ireland. Furthermore, this publication has emanated from research conducted with the financial support of Science Foundation Ireland (SFI) and the Department of Agriculture, Food and Marine on behalf of the Government of Ireland under Grant Number **16/RC/3835**.

Table 2: Percentages of Responses Made for Various Measures, by Material, in Study 1  (N=127 responses per material)

| Material | % Top-3 Answer Cats. | No. of Answer Cats. | % Goal Obj. | % Non-Goal Obj. | % Both Obj. | % Neither Obj. | % S1 Goal Objs | % S2 Non-Goal Objs | % S3 Goal Objs |
|---|---|---|---|---|---|---|---|---|---|
| alan_plane | 55% | 12 | 64% | 7% | 6% | 24% | 38% | 11% | 29% |
| anna_interview | 74% | 10 | 71% | 6% | 9% | 15% | 33% | 12% | 43% |
| belinda_meeting | 58% | 11 | 37% | 28% | 20% | 15% | 21% | 32% | 37% |
| bill_holiday | 49% | 12 | 48% | 9% | 2% | 40% | 23% | 19% | 58% |
| bob_job | 81% | 8 | 78% | 0% | 0% | 22% | 19% | 0% | 63% |
| edith_exam | 50% | 11 | 44% | 20% | 11% | 25% | 31% | 20% | 33% |
| john_party | 67% | 12 | 62% | 1% | 0% | 37% | 34% | 1% | 34% |
| karen_bus | 57% | 10 | 81% | 2% | 0% | 17% | 31% | 1% | 54% |
| mary_food | 69% | 8 | 80% | 6% | 2% | 13% | 64% | 1% | 29% |
| katie_kitten | 50% | 11 | 91% | 1% | 7% | 1% | 16% | 11% | 66% |
| louise_shopping | 63% | 10 | 64% | 7% | 6% | 24% | 21% | 2% | 68% |
| lucy_loan | 89% | 6 | 87% | 0% | 2% | 11% | 3% | 7% | 77% |
| michael_tea | 56% | 12 | 34% | 35% | 6% | 24% | 9% | 38% | 30% |
| peter_college | 56% | 12 | 59% | 2% | 1% | 38% | 37% | 3% | 27% |
| rebecca_swimming | 54% | 12 | 46% | 6% | 6% | 43% | 36% | 8% | 25% |
| robert_essay | 54% | 13 | 27% | 41% | 9% | 23% | 18% | 38% | 27% |
| sally_wine | 54% | 12 | 62% | 9% | 8% | 21% | 35% | 13% | 36% |
| sean_call | 65% | 10 | 61% | 10% | 6% | 24% | 31% | 9% | 44% |
| sam_driving | 61% | 11 | 64% | 5% | 9% | 23% | 31% | 14% | 35% |
| steve_gardening | 55% | 11 | 59% | 9% | 13% | 18% | 8% | 20% | 63% |